\documentclass[12pt]{article}
\usepackage{arxiv}
\usepackage[utf8]{inputenc} 
\pdfoutput=1
\usepackage[T1]{fontenc}    
\usepackage{hyperref}       
\usepackage{url}            
\usepackage{booktabs}       
\usepackage{amsfonts}       
\usepackage{nicefrac}       
\usepackage{microtype}      
\usepackage{lipsum}		
\usepackage{graphicx}
\usepackage{natbib}
\usepackage{doi}
\usepackage{amsmath}
\usepackage{algpseudocode}
\usepackage[ruled,linesnumbered]{algorithm2e}
\usepackage{multirow}
\usepackage{graphicx}
\usepackage{float}
\usepackage{subfigure}
\usepackage{setspace}

\title{Neural Theorem Provers Delineating Search Area Using RNN}

\author{Yu-hao Wu \\
	School of Mathematical Sciences\\
	University of Electronic Science and Technology of China\\
	Chengdu, China \\
	\And
	Hou-biao Li \\
	School of Mathematical Sciences\\
	University of Electronic Science and Technology of China\\
	Chengdu, China \\
}

\renewcommand{\shorttitle}{\textit{arXiv} Template}

\hypersetup{
pdftitle={Neural Theorem Provers Delineating Search Area Using RNN},
pdfsubject={q-bio.NC, q-bio.QM},
pdfauthor={Yuhao Wu, Houbiao Li},
pdfkeywords={First keyword, Second keyword, More},
}

\begin{document}
\maketitle

\begin{abstract}
Although traditional symbolic reasoning methods are highly interpretable, their application in knowledge graphs link prediction has been limited due to their computational inefficiency. A new RNNNTP method is proposed in this paper, using a generalized EM-based approach to continuously improve the computational efficiency of Neural Theorem Provers(NTPs). The RNNNTP is divided into relation generator and predictor. The relation generator is trained effectively and interpretably, so that the whole model can be carried out according to the development of the training, and the computational efficiency is also greatly improved. In all four data-sets, this method shows competitive performance on the link prediction task relative to traditional methods as well as one of the current strong competitive methods.
\end{abstract}

\keywords{Knowledge Graph \and Link Prediction \and Neural Symbolic Reasoning \and Deep Learning}

\section{Introduction}

Knowledge graphs (\cite{ehrlinger2016towards}) contains lots of real-world facts, which are useful in various applications. Each fact is typically specified as a triplet $(h, r, t)$
or another form $r(h, t)$, meaning entity $h$ has relation $r$ with entity $t$. For example, $father(Anakin SkyWalker, Luke SkyWalker)$ can express a fact that Anakin is Luke's father if
we have known the famous movie Star Wars. As above, still the example of Star Wars, when Luke knew that another character Leia Organa Solo
in this play is his sister? It is when Leia knew that her father is Anakin SkyWalker, too.
Even in the real world, people always can not know all the relations between things and things, people and people. As it is impossible to collect all facts, knowledge graphs are incomplete. Therefore,
a fundamental problem on knowledge graphs is to complete Knowledge graphs by reasoning with existing ones, also known as knowledge graphs reasoning.

Prevailing knowledge graphs reasoning use neural models (\cite{bordes2013translating,
lin2015learning,sun2019rotate,nickel2011three,yang2014embedding,nickel2016holographic,
trouillon2016complex,dettmers2018convolutional}) and symbolic reasoning models(
\cite{galarraga2013amie,galarraga2015fast,omran2018scalable,ho2018rule,niu2020rule}
), which
faced with kinds of problems including weak generalisation
results on datasets, discrete results which is unstable and hard to train with lots of modern optimization methods, lacking of explanatory which
will cause the model hard to improve its performance with the help of real-world experts.

\textbf{Neuro-Symbolic Reasoning} A promising direction for overcoming these issues consists in combining neural models and symbolic reasoning given their complementary strengths and weaknesses. Neuro-Symbolic (Neural Symbolic) Reasoning (\cite{guo2016jointly,
guo2018knowledge,zhang2019iteratively,richardson2006markov,qu2019probabilistic,
de2007problog,das2017go,yang2017differentiable}) performs well because
it contains explainable rules as well as it have continuous solution space.
As experts can play roles on explainable rules and modern optimization methods
can be used in continuous solution space, we focus on NTPs(\cite{rocktaschel2017end,
minervini2020differentiable,minervini2020learning}), a family of neuro-symbolic reasoning
models: NTPs are continuous relaxations of the backward-chaining reasoning
algorithm that replace discrete symbols with their continuous embedding representations.
NTPs have interesting properties: they can jointly learn representations and interpretable rules from data via back- propagation, and can potentially combine such rules in ways that
may have not been observed during training. However, a major limitation in NTPs is that, during training, they need to consider all rules for explaining a given goal or sub-goal. This quickly renders them ineffective in settings requiring a large number of rules or reasoning steps.

As Knowledge Bases(KBs) increasing, NTPs will generate a large number of sub-goals, which exponential growth base on the searching deep and KB's scale. It cause huge computational complexity and make NTPs can not solve problems with large datasets. Researcher make a lot of effort
to make NTPs easier to use and reduce the amount of computation. GNTPs(\cite{minervini2020differentiable}) dynamically constructing the computation graph of NTPs and including only the most promising proof paths during inference, thus obtaining orders of magnitude more efficient models. CTPs(\cite{minervini2020learning}) learn an adaptive strategy for
selecting subsets of rules to consider at each step of the reasoning process. This is
achieved by a select module that, given a goal, produce the rules needed for proving it.
Predicates and constants in the produced rules lie in a continuous embedding space.
Hence, the $select$ module is end-to-end differentiable, and can be trained jointly with the other modules via gradient-based optimisation. But their strategy to zoom out of the
search space have the disadvantage that the $select$ module is unexplained and hard to enhance with the help of experts. These methods improve the behavior of NTPs by adding
neural network layer or parameter, which is lack of ability to enhance it  using domain knowledge.

Our work focus on reducing the computational complexity of NTPs as well as making $select$ module controllable and explainable. Rnnlogic (\cite{qu2020rnnlogic}) provided an EM
algorithm(\cite{do2008expectation}) based
rule generator optimizer with explicit probability distribution representations. It use H values(a rule evaluation metric related to its predictor structure) to train rule generator. Together with the weight, it is cleared when the rule is assigned to the predictor. It is re-assigned in the e-step and passed into the generator later, which is equivalent to once the H value is used for
the rule generation once. Our approach combine with NTPs and use its feature that $and$ module's recursive searching ability may multiple access to one rule. So our approach make the score of a rule super-imposable and reduce the amount of calculation
per iteration under the use of GRU(\cite{chung2014empirical}), a kind of RNN(\cite{rumelhart1986learning}) network, which have fewer parameters
and semantic memory ability. $and$ module need the semantic memory ability to
generate more appropriate rules. Super-imposable rules scores is used to make
$and$ module better score the rules based on the goal's score.

\section{Related Work}
\subsection{End To End Differentiable Provers}
NTPs(\cite{rocktaschel2017end}) and its conditional proving strategies optimised version
CTPs(\cite{minervini2020learning}) are continuous
relaxation of the backward chaining algorithm: these algorithms works backward
from the goal, chaining through rules to find known facts supporting the proof.

Given a query(or goal) G, backward chaining first attempts to $unify$ it with the fact available in a given KB. If no matching fact is available, it considers all rules $H :- B$(We see facts as rules with no body and valuables), where $H$ denotes the head and $B$ the body, and $H$ can be unified with the query G resulting in a substitution for the variables contained in $H$. Then, the
backward chaining algorithm applies the substitution to the body $B$, and recursively attempts to prove until find the facts or catch the deep we have set.

Backward chaining can been seen as a type if $and/or$ search: $or$ means that any rule in the KB can be used to prove the goal, and $and$ means that all the premise of a rule must be proven recursively.

\textbf{Unification Module.}
In the backward chaining reasoning algorithm, $unification$
matches two logic atoms, such as $fatherOf(Anakin, Luke)$ and $dadOf(X, Y)$. It is backward chining reasoning's key operator, which play roles in discrete space.
In discrete spaces, equality between two atoms (e.g. $fatherOf \neq dadOf$ ) is evaluated
by $unification$ by examining the elements that compose them, and using substitution sets
(e.g. ${X/Anakin, Y/Luke} $) binds variables to symbols. In NTP, to be able to match different
symbols with similar semantics, $unification$ uses a Gaussian kernel to compare the similarity
of different representations in the embedding space.

In NTP, $unify_\theta (H, G, S) = S' $ generate a neural network$-$a proof state
$S = (S_\psi , S_\rho )$ consisting of a set of substitutions $S_\psi$ and a proof score
$S_\rho$. For example, given a goal $G = [fatherOf, Anakin, Luke]$ and a fact $H = [dadOf, X, Y]$,the $unify$ module uses Gaussian kernel $k$ to compare the embedding representations of
$fatherOf$ and $dadOf$, updates the variable binding
substitution set $S_\psi'=S_\psi\cup\{{X/Anakin, Y/Luke}\}$, and calculates the new proof score $S_\rho'= min(S_\rho,k(\theta_fatherOf,\theta_dadOf))$ and proof state $S'=(S_\psi',S_\rho')$.

\textbf{OR Module.}
The $or$ module traverses a KB, computes the unification between goal and all facts and rule heads in it, and then recursively use the $and$ module on the corresponding rule
bodies. Given a goal $G$ and each rule $H :- B$ with the rule head $H$ in
a KB $\mathfrak{K} $, module $or_\theta^\mathfrak{K} (G, d, S)$ unifies the goal $G$ with the rule head $H$, and bodies $B$ of each rule will be proved by using module $and$ until reach the set deepest depth $d$
or $unify$ fail. $or$ module is shown below:
\begin{equation}
    \begin{split}
        or_\theta^\mathfrak{K} (G, d, S) &= [S'|H:-B \in \mathfrak{K} ,\\
        &S' \in and_\theta^\mathfrak{K} (B, d, unify_\theta(h, g, s))]
    \end{split}
\end{equation}
For example, given a goal$ G = [grandpaOf, Q, Luke]$ and a rule $H :– B$ with $H = [grandfatherOf, X, Y] $
and $B = [[fatherOf, X, Z], [fatherOf, Z, Y]]$, $unyfy$ module compute the similarity of goal $G$ and the rule head $H$ to get a score, then $and$ module prove the sub-goals generated by rule body $B$ to get sub-scores.

\textbf{AND Module.}
After unification in $or$ module, $and$ module proves a list of sub-goals in a rule body $B$.
$and_theta^\mathfrak{k}(B:\mathbb{B},d,S)$ module first substitute variables in the first sub-goal B with constants
using substitutions in $S$, then use $or$ module to generate another sub-goals of B. $\mathbb{B}$ use the result state of above to prove the atoms by using the $and$ module recursively:
\begin{equation}
    \begin{split}
        and_\theta^\mathfrak{K} (B : \mathcal{B} , d, S) = &[S'' | d > 0, \\
        &S'' \in and_\theta^\mathfrak{K} (\mathcal{B} , d, S'), \\
        &S' \in or_\theta^\mathfrak{K} (sub(B, S_\phi), d - 1, S)]
    \end{split}
\end{equation}
For example, we use $and$ module to prove the rule body $B$ mentioned above.
$and$ module substitute variables with constants using substitutions in $S$ for the sub-goal $[fatherOf, X, Z]$,
then use the $or$ module to get a resulting state. $and$ module will be used to prove $[fatherOf, Z, Y]$ using resulting state generated above.

\textbf{Proof Aggregation.}
In a KB $\mathfrak{k}$, we use modules above to generate a neural network, which evaluates all the
possible proofs of a goal $G$. The largest proof score will be selected by NTPs:
\begin{equation}
    \begin{split}
        ntp_\theta^\mathfrak{K} (G, d)=max_SS_\rho \\
        with S \in or_\theta^\mathfrak{K} (G,d,(\emptyset,1))
    \end{split}
\end{equation}
where $d \in \mathbb{N}$, and $\mathbb{N}$ is defined at the beginning.
The initial proof state is set to $(\emptyset,1)$ which express an empty
substitution set and a proof score of 1.

\textbf{Training.}
NTPs minimise a cross-entropy loss $\mathcal{L}^\mathfrak{K} (\theta)$
to learn embedding representations of atoms using the final proof score.
Prover masking facts iteratively and try to prove them using other facts and rules.

\subsection{Deficiencies and improvement goals}
NTPs and its conditional proving strategies optimised version CTPs
In the NTPs proving process, evaluating effect of rules is complex due to the recursively
scoring process of $or$ module and $and$ module. CTPs reformulate goals by using extra neural network layers so that its training is end to end and easy to train.
But we can't know why a goal'relation can be reformulated due to its ambiguous manifestations.
We can't add new knowledge to interpret its relationship selection and optimize it.

In Rnnlogic(\cite{qu2020rnnlogic}), they provided a rule generator to generate rules, and use predictor to score rules by adding
H values to each rule if it path arrive at the destination:
\begin{equation}
    \begin{split}
        H(rule) = \{score_\omega (t|rule) - \frac{1}{|\mathcal{A}|} \sum_{e\in\mathcal{A}}
        score_\omega(e|rule)\}
    \end{split}
\end{equation}
where $\mathcal{A}$ is the set of all candidate answers discovered by rules. It leak the
ability to evaluate the importance of rules used by NTPs $and$ module's recursively
working process.Relation's contribution can't be evaluate correctly because the $H(rule)$ only
consider the distinctions which rules are used. The hierarchical proving process need Generator knows
that not all relations used will get the same treatment the next time they are generated, the deeper the relations should be generated, the greater the number.

We propose a model that not only makes the relation $selecte$ part interpretable, but also enables
relation selectors to better function in the hierarchical proof process of NTPs. We will introduce it in the
next chapter and demonstrate it experimentally in chapter four.

\section{EM-like Optimization with Hierarchical Relation Generator}
\begin{figure}[H]
	\centering
	\includegraphics[scale = 0.55]{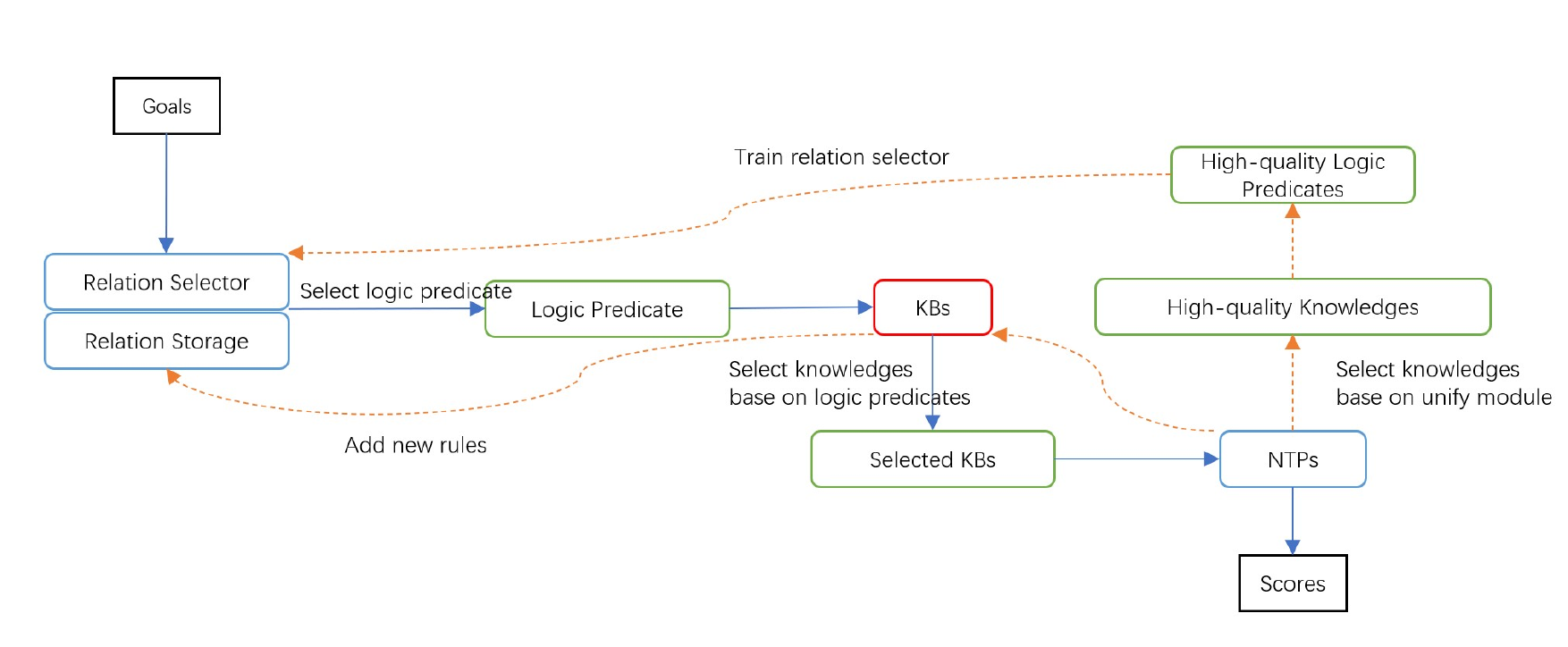}
	\caption{We proposed a hierarchical relation storage structure as well as a hierarchical Knowledge
	storage structure. They can store some knowledge hierarchically in the process of UNIFY,
	and after a predictor training is completed, the stored knowledge can be converted into some
	relations through the Nearest Neighbor Search(NNS) algorithm, and those relations will be used in the training of the
	relation generator.}
	\label{lct}
\end{figure}
In this section, we introduce our approach which learns logic rules for knowledge graph reasoning.
Considering NTPs have a recursively proving process, we provided a rule formulator
with a super-imposable rule scoring structure. We use an EM-like algorithm with implicit probability. First, we use rule formulator to generate
some rules.And in the e-step, we use these rules to train NTPs and recursively score the rules.
In the m-step, we use the scores of rules to train rule formulator.

\subsection{Relation Generator}
Considering that the proof process of NTPs is from top to bottom, layer by layer deeper,
there is a sequential relationship between the upper-level rule facts and the lower-level
rule facts, so this paper uses RNN as the rule and fact selector for selection. In the
proof process of the backward chain, if $unify$ succeeds in the proof process, then the
rules and facts traversed by the $or$ operation are the same as the previously traversed
rules and facts are related, and the related relationship is sequential. For example,
$grandpaOf\leftarrow fatherOf, parentOf$, when we want to prove the goal with the
relation $grandpaOf$, when we traverse to the above rule, we will expand $grandpaOf$ into two subgoals, each with $fatherOf The relationship between $ and $parentOf$; and when we want
to prove a goal with $fatherOf$, we hardly ever use a rule like
$fatherOf\leftarrow sonOf,grandparentOf$ to do a proof expansion. Therefore, the
sequence relationship between our upper-level rule facts and lower-level rule facts
is universal, and the use of RNN with sequence features for prediction can fully capture
the implicit relationship.We also call the relation generator $RNN_\theta$ below.

\textbf{Relation Selector:}
We propose an RNN-based relation selector to generate relation
sets according to the goals.First, relation selector has the generate structure $R-|B$,
$R$ is goal's relation(predicate) and $B$ is generated relation(predicate) consist of
$r_1,r_2,\dots,r_n$. We use a relation selector $RNN_\theta$ to predict sequence relations.
Given a goal sets $G$, we extract the relations $r$, Initialize $RNN_\theta$:
\begin{equation*}
	h_0=f(r),
\end{equation*}
The GRU gating unit is then used to make predictions on the rules at different locations:
\begin{equation*}
	h_t=GRU(h_{t-1},g([r_{t-1},r_t])),
\end{equation*}
$g$ is a linear transformation, $[r_{t-1},r_t]$is a connection between the previous relation and the current relation.

Perform a probability division on the generated rule set $R$ by $softmax(o(h_{t+1}))$ to
get the next relation $r_{t+1}$. Each relation $r_i$ is added to a relation set $Logic Predicates$,

\textbf{Relation Storage:}
Effectively training relation selectors is also a non-trivial task, so we propose a relational storage structure to better capture relational information from the predictor.
Using a sets of coefficient of expansion $\{ep_1, ep_2,\dots, ep_{maxDepth}\}$ related to the depth of the relations, we store relations like below:
\begin{equation}
	RelStorage = \{ \overbrace{r_1, \cdots, r_j}^{ep_1}, \overbrace{r_{j+1}, \cdots, r_{k}}^{ep_2*(j-1+1)}, \overbrace{r_{k+1}, \cdots, r_l}^{ep_3*(k-j)}, \cdots \}
\end{equation}
After some processing, we use $RelStorage$ as the training set of Relation Selector, and the details are explained
in the following training process.

\subsection{Predictor base on NTPs}
We remodeled some modules of NTPs to evaluate relations
and accelerate computing.
	
\begin{algorithm}[H]
	  \caption{Or Module}
	  \label{al or}
	  \textbf{function} or(G, d, S):\\
	  \For{$H :- \mathbb{B} \in Selected KBs$}{
		  \For{$S\in and(\mathbb{B},d,unify(H,G,S))$}{
			  yield S
		  }
	  }
	\end{algorithm}
	\begin{algorithm}[H]
	  \caption{Unify Module}
	  \label{al unify}
		\textbf{function} $unify(H,G,S=(S_\psi ,S_\rho ))$:\\
		$S_\psi'=S_\rho\bigcup_i T_t $\\
		with $T_i = \left\{
			\begin{aligned}
			\{H_i/G_i\} & if H_i \in \mathcal{V} \\
			\{G_i/H_i\} & if G_i \in \mathcal{V},H_i \notin \mathcal{V} \\
			\emptyset & otherwise
			\end{aligned}
			\right.$\\
			\eIf{$S_\rho<setedMinScore$}{\textbf{break}}(){High-quality Knowledge \textbf{add} H}
		$S_\rho'=min\{S_\rho\}\bigcup_{H_i,G_i\notin \mathcal{V}}\{K(\theta_{H_i},\theta_{G_i})$\\
		\textbf{return} $(S_\psi' ,S_\rho' )$
	\end{algorithm}

\textbf{Or Module.}NTPs traverse all knowledge in KBs in the process of proof, which cause some computational issues. We reformulated it and make it only consider the knowledge we provide.
As algorithm\ref{al or} shown, we only traverse the knowledge matched in the KBs by the relations
generated by the generator. Other parts are consistent with NTPs.

\textbf{Unify Module.} In NTPs, unify module returns a new substitution set and a new goal's score base on the similarity of relation and entity. As algorithm\ref{al unify} shown, we set a threshold
$setedMinScore$ to control the unify module. If the similarity exceeds a threshold, we add this knowledge
to our high-quality level for subsequent operations. And if not, terminate the proof branch.

\subsection{Model Training}

As figure \ref{lct} show, a hierarchical relation storage structure and a hierarchical Knowledge
storage structure a provided to prepare the training data for the generator. Generator and predictor
are trained together by a EM-like algorithm.

\begin{algorithm}[H]
	\caption{Basic traning process}
	\label{alg:algorithm1}
	\KwIn{triple data set goals $[(h,r,t)]$, number of iterations $n$.}
	\KwOut{Scores, trained generator $RNN_\theta$, trained predictor $NTPs_\theta$.}
	\BlankLine
	Initialize KB, $RNN_\theta$, $NTPs_\theta$;
	
	\For{i = 1 to $n$}{
		$RNN_\theta$ generate Logic Predicates\;
		Selected KBs = Select(KBs, Logic Predicates)\;
		\textbf{e-step:} start train $NTPs_\theta$:\\
		
		\If{unify(facts or rules$\in Selected KBs$) not FAIL}{
			High-quality Knowledge add (facts or rules);
		}
		\While{size(Relation Storage) < maxSize(Relation Storage)}{
			High-quality Knowledge add NNS(KB)\;
			Relation Storage add predicates of High-quality Knowledge\;
		}
		\textbf{m-step:} train Relation Selector with Relation Storage;
	}
\end{algorithm}

Algorithm \ref{alg:algorithm1} shows the basic process of training. We first
initialize the generator $RNN_\theta$ and predictor $NTPs_\theta$. The relations of this batch of goals are input to the generator, and then the generator is used to generate a series of relations;
The generated relations are matched in KBs to generate a knowledge domain. In the e-step, our predictor uses the knowledge in this domain to score this batch of goals, and put the useful knowledge into the $High-quality knowledge$.
When the size of $Relation Storage$ does not reach the set value,
we need to use NNS to search for $High-quality Knowledge$ in KB first, and add the predicate(relation) of $High-quality Knowledge$ to $Relation Storage$.
In the m-step, we connection the values in the $Relation Storage$ with the goals' predicates as training data to train the selector.

\textbf{KB initialization:} Read in the triplet data set, and create a tuple with the relationship or rule header as the key and the entity pair or rule body as the value for easy access.

\textbf{Generator initialization:} Establish an RNN network based on GRU gates, where the dimension of $h_t$ is consistent with the relation's embedding size.

\textbf{Predictor initialization:} Because the embedding structure of the complex matches the $unify$ operation in NTPs, we first use the complex to perform a pre-training on all KB data.

\textbf{Generate knowledge set $Selected KBs$:} $RNN_\theta$ traverses each relation in the goal, generates a series of relation sets $Logic Predicates$, and performs each relationship in
$KB$ for each relationship in $Logic Predicates$ matching, matching rules and facts generate a knowledge set $Selected KBs$.

\textbf{Training predictor $NTPs_\theta$:} NTPs are the same as described above, but the $OR$ module is modified,
and the current facts or rules are added to $High-quality Knowledge$ every time $unify$ succeeds. The other parts of the algorithm are basically the same as NTPs described above.

\textbf{High-quality Knowledge completion:} Use Nearest Neighbor Search(NNS)(\cite{tao2002continuous}) domain search algorithm to search the knowledge in High-quality Knowledge in $KB$, add the closest knowledge to Relation Storage until the set upper limit is reached.

\section{Experiment}

\subsection{Experiment Settings}

\textbf{Datasets:Kinship, UMLS, Nations and Countries.} In our experiment, we choose three benchmark datasets for
evaluation, which are Alyawarra kinship(Kinship), Unified Medical Language System(UMLS), Nations.
For the Kinship, UMLS and nations datasets.Nations contains 56 binary predicates, 111
unary predicates, 14 constants and 2565 true facts; kinship contains 26 predicates, 104 constants and 10686 true facts;
UMLS contains 49 predicates, 135 constants and 6529 true facts real facts. Because our benchmark ComplEx cannot
handle unary predicates, we remove unary atoms from Nations. For the Kinship, UMLS and nations datasets, there are no standard data splits. So we split each knowledge base into 30\%
training facts, 20\% validation facts and 50\% testing facts. For evaluation, we take a test fact and
corrupt its first and second arguments in all possible ways such that the corrupted fact is not in the
original KB. Subsequently, we predict each test fact and its corrupted rank to compute \textbf{MRR} and HITS@m.
We also use the COUNTRIES dataset (Bouchard, Singh, and Trouillon 2015) to evaluate the
scalability of our algorithm. The dataset contains 272 constants, 2 predicates, and 1158 ground truths,
and is designed to explicitly test the logical rule induction and reasoning abilities of link prediction
models. We compare (Rockt¨aschel and Riedel 2017), inference steps that require increasing length and difficulty (S1, S2, S3).

\textbf{Evaluation Benchmark.} In Kinship, UMLS, Nations and Countries, we predict each test fact and its corrupted rank to compute \textbf{MRR} and \textbf{HITS@m}(m=1,3,10) after training. In country,
we evaluate the area under \textbf{Precision-Recall-curve (AUC-PR)} with results comparable to previous methods.
\textbf{Average training time per iteration(relatively)}(ATTP) is set to compare the computational
performance. \textbf{Knowledge Utilazition} show the utilization of knowledge every time we use, it
counting through the success of unify's branch establishment and comparing the knowledge this batch's goals
use. ATTP and Knowledge Utilization can evaluate the expected performance on large datasets.

\textbf{Modules Setting.} In relation generator, we use embedding that comes with pytorch. And in predictor,
we use ComplEx to pretrained the datasets because it fits the way $unify$ works.
The scale proportion of $Selected KBs$ is set to 30\%. The number of  $Relation Storage$ layers is the same
as the number of recursive layers of NTPs, set to 3.

\subsection{Results and Analysis.}
\textbf{Accuracy.} Compared to other neuro-symbolic inference models, we get a good performance at HITS@10, especially
in Kinship. It shows that we can easily narrow down the correct result to a small range.
Because the calculation time and the amount of traversal knowledge are too small compared to NTP,
NeuralLp, MINERVA, our final accuracy has dropped.

\begin{table}[H]
	\renewcommand\arraystretch{1.5}
	\centering
	\caption{Results compared to NTPs and previous methods}
	\begin{tabular}{llllllll}
	\hline
	\multirow{2}{*}{Datasets}  &    & \multirow{2}{*}{Metrics} & \multicolumn{5}{c}{Models}                                                             \\
	\cline{4-8}
							   &    &                          & Ours  & CTP   & NTP                                              & NeuralLP & MINERVA  \\
	\hline
	\multirow{4}{*}{Nations}   &    & MRR                      & 0.701 & 0.709 & 0.74                                             & -        & -        \\
							   &    & HITS@1                   & 0.539 & 0.562 & 0.59                                             & -        & -        \\
							   &    & HITS@3                   & 0.875 & 0.813 & 0.89                                             & -        & -        \\
							   &    & HITS@10                  & 0.997 & 0.995 & 0.99                                             & -        & -        \\
	\hline
	\multirow{4}{*}{Kinship}   &    & MRR                      & 0.772 & 0.764 & 0.80                                             & 0.619    & 0.720    \\
							   &    & HITS@1                   & 0.609 & 0.646 & 0.76                                             & 0.475    & 0.605    \\
							   &    & HITS@3                   & 0.891 & 0.859 & 0.82                                             & 0.707    & 0.812    \\
							   &    & HITS@10                  & 0.973 & 0.958 & 0.89                                             & 0.912    & 0.924    \\
	\hline
	\multirow{4}{*}{UMLS}      &    & MRR                      & 0.801 & 0.852 & 0.93                                             & 0.778    & 0.825    \\
							   &    & HITS@1                   & 0.589 & 0.752 & 0.87                                             & 0.643    & 0.728    \\
							   &    & HITS@3                   & 0.951 & 0.947 & 0.98                                             & 0.869    & 0.900    \\
							   &    & HITS@10                  & 0.972 & 0.984 & 1.00                                             & 0.962    & 0.968    \\
	\hline
	\multirow{3}{*}{Countries} & S1 & \multirow{3}{*}{AUC-PR}  & 100.0 & 100.0 & \begin{tabular}[c]{@{}l@{}}100.00\\\end{tabular} & 100.00   & 100.0    \\
							   & S2 &                          & 89.47 & 91.81 & 93.04                                            & 75.1     & 92.36    \\
							   & S3 &                          & 95.21 & 94.78 & 77.26                                            & 92.20    & 95.10    \\
	\hline
	\end{tabular}
\end{table}
\begin{figure}[H]
	\centering
	\includegraphics[scale = 0.55]{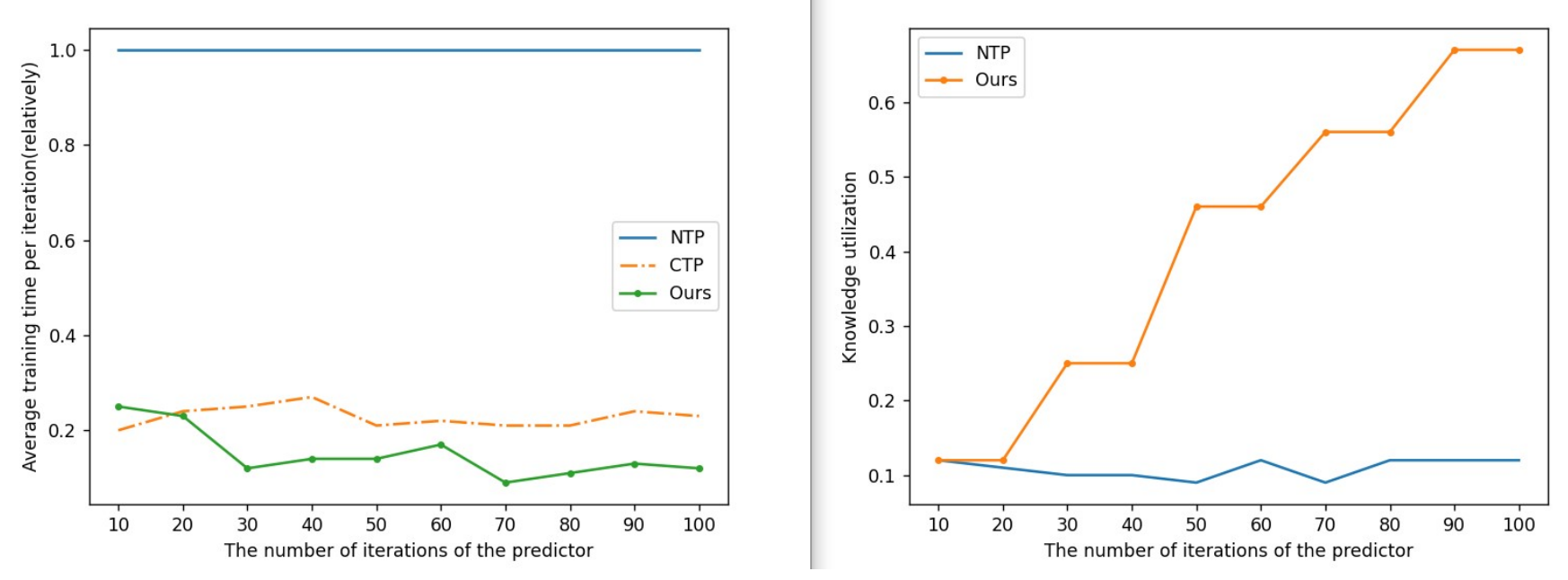}
	\caption{Knowledge utilization and average training time(relatively) per iteration compared to
	NTP and CTP. We can observe that our model has a performance advantage over previous models.
	(The knowledge utilization of CTP is hard to evaluate.Based on the fact that it improves performance
	by reformulating the goals, we cannot properly evaluate its performance by knowledge utilization. )}
	\label{xgt}
\end{figure}

\begin{figure}[H]
	\centering
	\includegraphics[scale = 0.57]{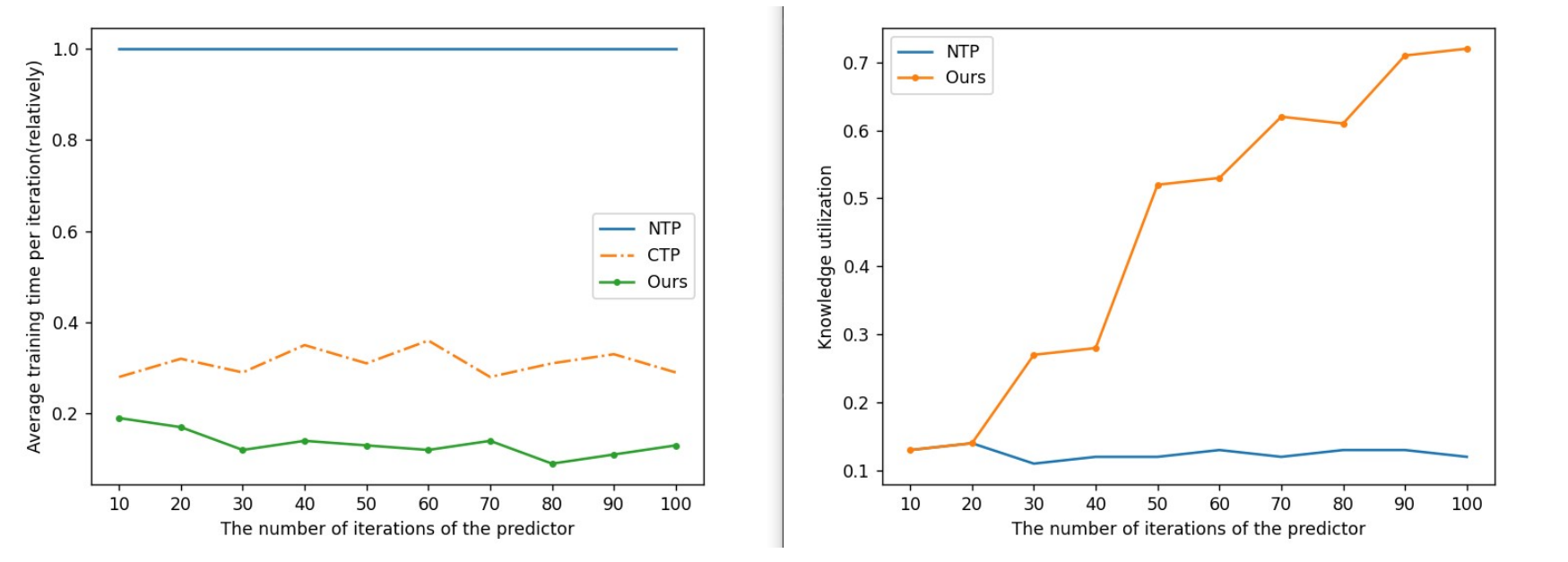}
	\caption{In Nations dataset, 2-hop and 1-hop reverse reasoning significant performance drain in NTPs.
	Our method limits the scope of knowledge search, and performance gains are relatively large in these tasks.}
	\label{nations}
\end{figure}

\textbf{Computing Performance}. As figure \ref{xgt} shown, ATTP and Knowledge Utilazition have a huge improvement over NTP and a relatively small improvement compared to CTP.
Especially in Nations dataset, as figure \ref{nations} shown, which contains lots of rules with 2 body atoms.
Our method is better than NTP in performance, because in multi-hop reasoning, NTP's search method can cause serious performance bottlenecks, while our method makes the search domain learnable and the search range is smaller each time .
Considering that the accuracy is not much different from CTP, the improvement in our computational performance is encouraging.

Compared with the most traditional symbolic reasoning methods, the accuracy has decreased,
but our performance has increased by more than ten times, which is acceptable, and CTP has also
set an example in this regard. Compared with CTP, our other advantage is that we can manually control
the training of the generator by adjusting $Relation Storage$, which means that we can also use human
strength to make the training results better.

For example, for a known specific task, such as related to the word $grandfather$, we can train in advance
by adding words such as $father$ and $son$ to the sequence backbend of the warehouse to get a better
acceleration effect. In the EM algorithm, this is equivalent to performing a fine-tuning of
the implicit distribution of the intermediate variables in advance.

\section{Conclusion and Future Work}
Using neural symbolic reasoning method to reason in knowledge graph has the advantage of strong
interpretability, and it is easy to optimize according to new knowledge.
Our method corrects the shortcomings of low computational efficiency of
traditional methods and the weak interpretability of some modules of CTP.
NTP has good interpretability and can generate new rules. After our method improves its
computational efficiency, it can be used more widely. Our method is more efficient, so that such method can be placed in more
complex knowledge graphs in the future.

However, we also have some shortcomings. Although the hierarchical structure
of the relationship corresponding to the backward chain reasoning is used to
construct the RNN, its training effect is limited compared to the CTP. Finding a
new and more efficient relational generator and a more efficient training structure
for constructing the generator will be my future focus in this model. This method based on
the EM algorithm needs to combine the structure of the predictor and the generator, and
can be used in more predictors in the future to improve their training efficiency and preferences.

\bibliographystyle{unsrtnat}
\bibliography{references}  






\end{document}